\documentclass[11pt]{article}
\PassOptionsToPackage{dvipsnames}{xcolor}
\usepackage[margin=1in]{geometry}
\usepackage[T1]{fontenc}
\usepackage{mathptmx}
\usepackage{microtype}
\usepackage{amsmath}
\usepackage{graphicx}
\usepackage{booktabs}
\usepackage{array}
\usepackage{tabularx}
\usepackage{multirow}
\usepackage{wasysym}
\usepackage{pifont}
\newcommand{\cmark}{\ding{51}}
\usepackage[edges]{forest}
\usepackage{enumitem}
\usepackage[font=small,labelfont=bf]{caption}
\usepackage[round,sort,authoryear]{natbib}
\usepackage[dvipsnames]{xcolor}
\usepackage[colorlinks=true,linkcolor=MidnightBlue,citecolor=MidnightBlue,urlcolor=MidnightBlue]{hyperref}
\let\cite\citep
\setlist{itemsep=0.2em,topsep=0.3em}

\title{\textbf{From Long Context to Selective Memory:\\A Survey of Memory-Centric Compression for Large Language Models}}
\author{Chaogui Gou\\[0.3em]
\small USTB\\
\small \texttt{ustbedwin@gmail.com}}
\date{}

\usepackage{accsupp}
\usepackage{letltxmacro}\LetLtxMacro\oldCIRCLE\CIRCLE\LetLtxMacro\oldLEFTcircle\LEFTcircle\LetLtxMacro\oldCircle\Circle\LetLtxMacro\oldemptyset\emptyset
\DeclareRobustCommand{\CIRCLE}{\BeginAccSupp{method=hex,unicode,ActualText=25CF}\oldCIRCLE\EndAccSupp{}}
\DeclareRobustCommand{\LEFTcircle}{\BeginAccSupp{method=hex,unicode,ActualText=25D0}\oldLEFTcircle\EndAccSupp{}}
\DeclareRobustCommand{\Circle}{\BeginAccSupp{method=hex,unicode,ActualText=25CB}\oldCircle\EndAccSupp{}}
\DeclareRobustCommand{\emptyset}{\mathord{\BeginAccSupp{method=hex,unicode,ActualText=2205}\oldemptyset\EndAccSupp{}}}

\begin{document}
\maketitle

\begin{abstract}
Long-context capability has become essential for large language models (LLMs) to handle complex tasks, yet continuously expanding the context window does not remove the computational, storage, and information-selection pressures incurred at inference time. Existing work on prompt compression, KV-cache management, efficient serving, retrieval augmentation, and agent memory is usually discussed in isolation, leaving the flow of information between storage and computation insufficiently understood. This survey reframes long-context compression as a memory-management problem for allocating inference-time resources. We distinguish compute-active memory, which directly participates in model computation, from compute-inactive memory, which lies outside the computation path but can be stored, retrieved, updated, or forgotten, and we characterize offloaded KV caches, persistent prefix caches, and precomputed KV states as intermediate states on the activation path. On this basis, we formulate writing, offloading, recall, retrieval-and-injection, materialization, eviction, consolidation, and forgetting as state transitions within an information lifecycle, and introduce a budgeted cost model that places these operations in a single cost account. The resulting framework unifies short-term context compression, long-term memory management, and the transitions between them, and shows that the central challenge is not merely reducing information volume but deciding which information should participate in costly inference, when, and in what form. We further summarize evaluation dimensions, benchmark gaps, and open challenges, and argue that future research should move from static, pointwise compression toward dynamic, governable, lifecycle-aware memory management.
\end{abstract}


\section{Introduction}

Long-context capability has become a prerequisite for large language models (LLMs) on complex tasks. As context windows have grown from several thousand to hundreds of thousands or even millions of tokens~\cite{deepseekai2026deepseekv4,openai2026gpt5.5,google2026gemini3.1pro}, long-document understanding, codebase analysis, multi-turn dialogue, tool-use trajectories, retrieval-augmented generation (RAG), and long-term personalized assistance increasingly depend on a model's ability to receive and exploit large amounts of contextual information~\cite{liu2025comprehensivelclm}.

A larger window, however, does not by itself yield more effective use of information~\cite{liu2024lost,hsieh2024ruler}. Long inputs increase prefill computation and place growing pressure on key-value (KV) cache memory, memory bandwidth, and latency during decoding~\cite{zhang2023h2o,kwon2023pagedattention,deepseekai2026deepseekv4}, while realistic contexts contain repeated, obsolete, or irrelevant content that dilutes critical evidence. The long-context problem is therefore not only one of capacity but also one of management: how a system selects, compresses, and uses information under a finite computational budget.

Several research directions address this problem at different levels, including prompt compression, latent memory representations, KV-cache compression, architecture-level selective computation, efficient serving, retrieval augmentation, and external long-term memory~\cite{jiang-etal-2023-llmlingua,jiang-etal-2024-longllmlingua,mu2023learning,zhang2023h2o,kwon2023pagedattention,lewis2020rag,packer2023memgpt}. Existing surveys usually organize them by technical origin or application scenario~\cite{liu2025comprehensivelclm,liu2025contextextension,verma2024contextualcompression}, which obscures a question they share: which information should enter the expensive attention-computation path, and which should instead be compressed, externalized, reused, updated, or forgotten?

This survey answers this question from a memory-centric perspective. We call information that directly participates in the current model computation \emph{compute-active memory}, and information that lies outside the current computation path but can be stored, retrieved, updated, or forgotten \emph{compute-inactive memory}. The distinction concerns an item's participation in computation at a given inference step rather than its storage location: the same content may be stored externally, later retrieved and injected into the context, and finally written back after inference. Between the two lie intermediate states, in which information has left the attention path but can be restored without re-retrieval or re-encoding, as with offloaded KV caches and precomputed prefix caches. Long-context compression thus becomes the management of an information lifecycle across these states. Recent surveys have also proposed unified views of LLM memory based on memory operations, context engineering, or agent memory~\cite{du2025rethinking,mei2025contextengineering,hu2025memoryage}; we instead organize the field by the computational participation of information and its cost. The main contributions are as follows.

\begin{enumerate}
\item We propose a framework that distinguishes compute-active from compute-inactive memory, incorporates intermediate states on the activation path, and casts long-context compression as lifecycle management under a finite computational budget.
\item We formulate writing, offloading, recall, retrieval-and-injection, materialization, eviction, consolidation, and forgetting as state-transition operators and introduce a budgeted cost model that places these operations in a single cost account.
\item We review representative work up to 2026 along this lifecycle, from input, representation, and KV-cache compression to external long-term memory, learned memory policies, and transition mechanisms such as retrieval-and-injection, KV offloading and materialization, and agent-side context folding.
\item We summarize evaluation dimensions, benchmark gaps, and open challenges, and pose a concrete research question for each challenge.
\end{enumerate}

\paragraph{Scope.} This survey focuses on the intersection of long-context modeling, KV-cache management, retrieval augmentation, and agent memory: how a system decides which information enters the current inference path, compresses its representation, controls its cost, and later updates or forgets it. Methods that only extend the context window without changing information selection, cache management, or memory mechanisms are treated as background. Parametric memory is included only in a limited way, because knowledge editing, continual pre-training, and parameter-efficient fine-tuning incur their cost mainly at training time and do not selectively activate information within a single inference; methods that blur this boundary, such as test-time neural memory~\cite{behrouz2025titans} and self-study-trained KV caches~\cite{eyuboglu2025cartridges}, are discussed where they affect inference-time memory allocation.

The remainder of the survey is organized as follows. Section~\ref{sec:background} reviews the costs of long-context inference and positions this survey relative to prior reviews. Section~\ref{sec:framework} presents the framework. Sections~\ref{sec:active} and~\ref{sec:inactive} review compute-active and compute-inactive memory, Section~\ref{sec:transition} discusses state transitions, Section~\ref{sec:evaluation} covers evaluation, Section~\ref{sec:challenges} discusses open challenges, and Section~\ref{sec:conclusion} concludes.

\section{Background and Positioning}
\label{sec:background}

\subsection{From Context Extension to Context Management}

Early work on long-context LLMs focused on enlarging the input that a model can accept, through positional-encoding extrapolation, sparse and sliding-window attention, recurrent-state modeling, retrieval augmentation, and long-context fine-tuning~\cite{chen2023positioninterpolation,peng2023yarn,ding2024longrope,beltagy2020longformer,zaheer2020bigbird,dai2019transformerxl,borgeaud2022retro,chen2024longlora}. These methods allow a model to access more complete task background in a single inference, which matters for long-document understanding, cross-file code analysis, multi-turn dialogue, and tool use, where relevant information is scattered across texts, knowledge bases, interaction histories, and execution states~\cite{bai2024longbench,liu2024repobench,lewis2020rag,yao2023react,zheng2024sglang,packer2023memgpt}. As windows reach hundreds of thousands or millions of tokens~\cite{ding2024longrope,deepseekai2026deepseekv4}, the central question has shifted from how much information a model can receive to which information deserves to enter the current inference path, and at what cost~\cite{liu2025contextextension,liu2025comprehensivelclm}.

This shift is driven by two kinds of cost. The first is computational. Prefill cost and time-to-first-token grow with input length; during decoding, the KV cache grows with context length, model size, and batch size and consumes GPU memory and bandwidth; and information kept in external indexes becomes costly again once it is retrieved and injected into the prompt. The second is informational. Long inputs mix key evidence with background, repetition, outdated states, and noise, and models do not use information uniformly across positions~\cite{liu2024lost}; when evidence is dispersed or appears in the middle of the input, omissions, evidence mismatch, and inconsistent reasoning become more likely. Long-context compression therefore differs from generic model acceleration: besides reducing computation, cache usage, and latency, it must preserve evidence coverage, fidelity, temporal consistency, and future recoverability. Over-compression breaks evidence chains, whereas over-retention increases latency, memory pressure, and noise.

\subsection{Existing Surveys and the Gap Addressed by This Survey}

Prior surveys fall into four groups. Surveys of long-context LLMs~\cite{liu2025comprehensivelclm,liu2025contextextension,liu2025thusspake} cover positional extension, sparse attention, training, infrastructure, and evaluation; compression is one topic among many, and transitions of information between internal states and external storage are rarely analyzed. Surveys of prompt compression~\cite{li2025promptcompression}, KV-cache management~\cite{li2024kvcachesurvey,xu2026kvstrategies}, and efficient serving~\cite{zhou2024efficientinference,miao2023efficientserving} focus on the cost of the current inference path, but give little attention to how long-term memory is written, organized, and forgotten, and seldom relate KV offloading to external-memory retrieval. Surveys of RAG and agent memory~\cite{gao2023ragsurvey,verma2024contextualcompression,zhang2024agentmemorysurvey,du2026autonomousmemory} emphasize external knowledge and long-term history; \citet{hu2025memoryage} organize agent memory by form, function, and dynamics; they include KV compression as latent working memory but assign model-internal cache management to LLM memory, outside the scope of agent memory. The survey of context compression for LLM agents by \citet{ctxcomp2026agents} is closest in spirit: it describes a select--compress--store--recover pipeline, discusses recoverability and control overhead, and formulates learned compression control as task reward minus cost. Its scope, however, is application-level trajectory compression, and it excludes KV-cache and model-level methods. Neither group distinguishes re-encoding retrieved text from recalling states that are already encoded, or accounts for online activation, offline materialization, cross-request reuse, and invalidation within one cost structure.

The fourth group, closest to this survey, proposes unified views of LLM memory. \citet{du2025rethinking} distinguish parametric from contextual memory and define six atomic operations across long-term memory, long context, and parametric modification; \citet{wu2025humanmemory}, \citet{shan2025cognitive}, and \citet{jia2026hippocampus} organize LLM memory by analogy to human memory; and the context-engineering survey of \citet{mei2025contextengineering} places memory hierarchies and context compression under context management. On the system side, MemGPT~\cite{packer2023memgpt} manages context through an operating-system analogy, and MemOS~\cite{li2025memos} schedules plaintext, activation, and parametric memory within one system. These works organize the field by cognitive analogy, functional role, or operation type.

This survey differs in three respects. First, it organizes methods by whether and when information participates in inference computation, which allows system-level mechanisms such as offloaded KV caches, persistent prefix caches, and offline-compressed KV to be compared with external memory within one lifecycle. Second, it treats state transitions as first-class objects of analysis, in particular the distinction between re-encoding retrieved text (retrieve-and-inject) and recalling states that are already encoded (recall), which fetch memory at very different costs. Third, it compares methods through a budgeted cost model that separates online, offline, and amortized costs, so that costs moved from the current context into background maintenance are not hidden. Table~\ref{tab:related-survey-comparison} summarizes the comparison.

\begin{table}[!tbp]
\centering
\scriptsize
\caption{Comparison with representative related surveys. \oldCIRCLE: systematically discussed; \oldLEFTcircle: partially discussed; \oldCircle: largely not discussed. Ratings reflect the authors' reading of each survey's structure and coverage.}
\label{tab:related-survey-comparison}
\setlength{\tabcolsep}{3pt}
\renewcommand{\arraystretch}{1.15}
\begin{tabularx}{\textwidth}{@{}>{\raggedright\arraybackslash}p{3.0cm}>{\raggedright\arraybackslash}p{2.7cm}*{4}{>{\centering\arraybackslash}p{1.15cm}}>{\raggedright\arraybackslash}X@{}}
\toprule
\textbf{Survey} & \textbf{Organizing perspective} & \textbf{KV / serving} & \textbf{Long-term / agent memory} & \textbf{Explicit cost model} & \textbf{State transition / injection} & \textbf{Relation to this survey} \\
\midrule
Long-context LLM surveys~\cite{liu2025comprehensivelclm,liu2025thusspake} & Architecture, infrastructure, training, evaluation & \CIRCLE & \Circle & \LEFTcircle & \Circle & Background on long-context modeling (Sec.~\ref{sec:background}) \\
Prompt compression~\cite{li2025promptcompression} & Hard / soft prompt compression & \Circle & \Circle & \LEFTcircle & \Circle & Admission control for active memory (Secs.~\ref{sec:prompt}--\ref{sec:latent}) \\
KV-cache management~\cite{li2024kvcachesurvey} & Token / model / system level & \CIRCLE & \Circle & \LEFTcircle & \Circle & System-level basis (Secs.~\ref{sec:kv}--\ref{sec:serving}, \ref{sec:intermediate}) \\
Agent memory mechanisms~\cite{zhang2024agentmemorysurvey} & Memory sources, forms, operations & \Circle & \CIRCLE & \Circle & \LEFTcircle & Writing and maintenance of external memory (Sec.~\ref{sec:inactive}) \\
From Human Memory to AI Memory~\cite{wu2025humanmemory} & Object $\times$ form $\times$ time & \LEFTcircle & \CIRCLE & \Circle & \Circle & Cognitive analogy; we adopt a cost view \\
Cognitive Memory in LLMs~\cite{shan2025cognitive} & Sensory / short-term / long-term & \LEFTcircle & \CIRCLE & \Circle & \LEFTcircle & Covers KV memory without modeling transition cost \\
Rethinking Memory in AI~\cite{du2025rethinking} & Parametric / contextual memory and six atomic operations & \LEFTcircle & \CIRCLE & \Circle & \LEFTcircle & Closest operation view; we define operators by state change (Sec.~\ref{sec:operators}) \\
Context engineering~\cite{mei2025contextengineering} & Retrieval, processing, management, and system implementations & \LEFTcircle & \CIRCLE & \Circle & \LEFTcircle & Broader scope; we focus on compression and activation cost \\
Memory in the Age of AI Agents~\cite{hu2025memoryage} & Forms $\times$ functions $\times$ dynamics & \LEFTcircle & \CIRCLE & \Circle & \LEFTcircle & KV compression as latent working memory; cache management left to LLM memory \\
Context compression for LLM agents~\cite{ctxcomp2026agents} & Target $\times$ mechanism $\times$ control policy & \Circle & \LEFTcircle & \LEFTcircle & \LEFTcircle & Application-level; excludes KV-level methods (Sec.~\ref{sec:agent-activation}) \\
\midrule
\textbf{This survey} & Compute-active / compute-inactive memory, lifecycle transitions, and a cost model & \CIRCLE & \CIRCLE & \CIRCLE & \CIRCLE & --- \\
\bottomrule
\end{tabularx}
\end{table}

\section{Memory-Centric Framework for Long-Context Compression}
\label{sec:framework}

\begin{figure}[!tbp]
\centering
\includegraphics[width=0.9\linewidth]{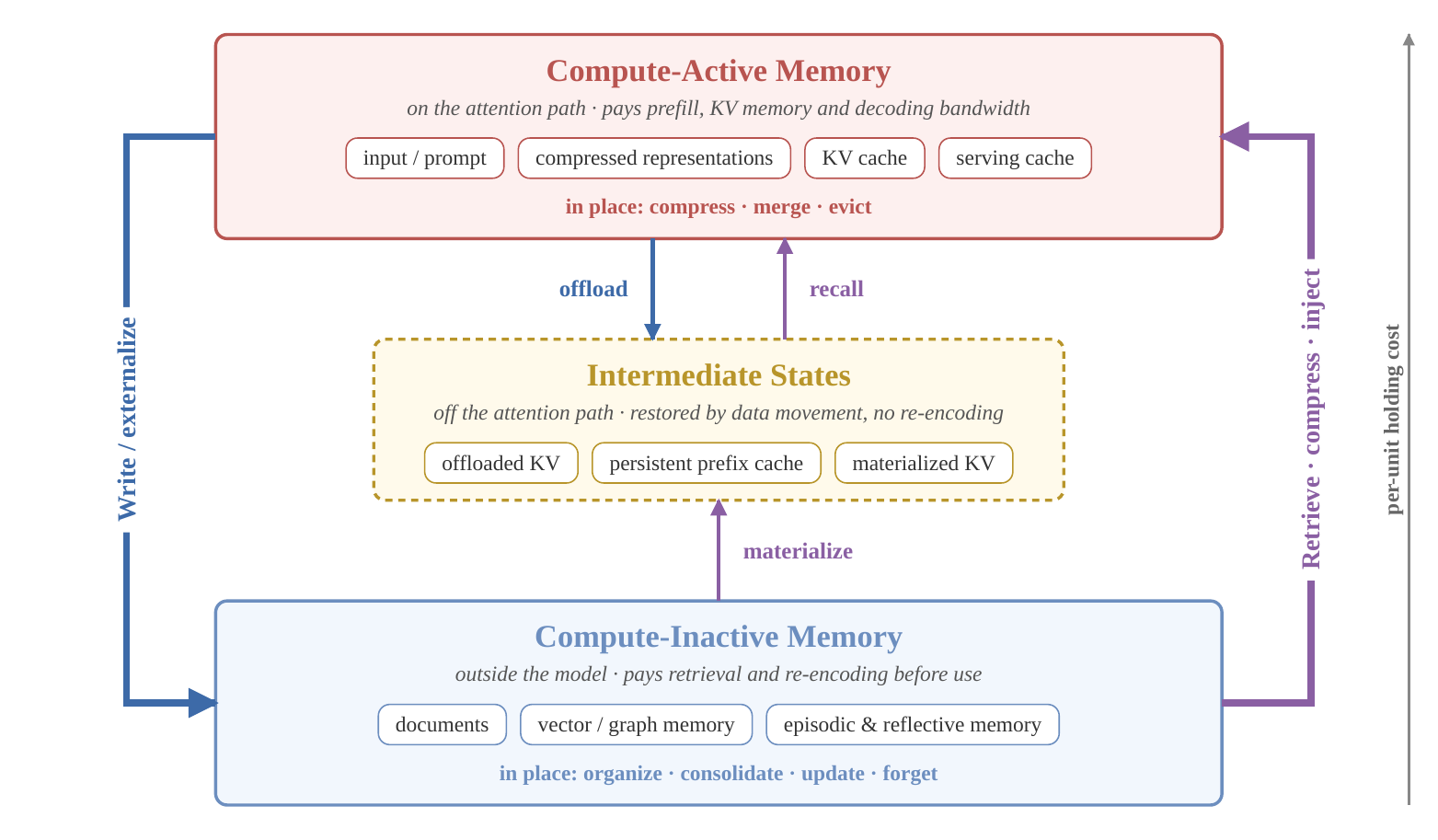}
\caption{Memory-centric lifecycle view of long-context compression. Information flows between compute-active and compute-inactive memory: writing externalizes information from the inference path, while retrieval, compression, and injection reactivate external memory. Offloaded KV caches, persistent prefix caches, and materialized KV states are intermediate states on the activation path that can be recalled directly. In-place operations compress, reorganize, or discard content within a state; per-unit holding cost increases from bottom to top.}
\label{fig:lifecycle}
\end{figure}

We treat long-context compression as a memory-management problem of allocating inference-time computation: the goal is to decide when information should participate in inference, in what form, and at what cost. This section defines memory states, the operators that move information between them, the resulting lifecycle, and a cost model that underlies the taxonomy of later sections. Figure~\ref{fig:lifecycle} gives an overview.

\subsection{Memory States and Computational Participation}

We distinguish memory states by participation in inference computation, because this criterion corresponds directly to the main costs of a long-context system. We fix its granularity at a single request: information is compute-active if it is encoded for the current model and addressable by attention during the request, that is, resident in accelerator memory as tokens, latent states, or KV caches. Such information incurs prefill computation, accelerator memory, memory bandwidth, and decoding latency. Methods that attend to only part of it at each decoding step, such as page-level sparse attention, reduce reads and attention computation within compute-active memory rather than moving information out of it.

\emph{Compute-active memory} is information that the model accesses during the current inference: prompt tokens, soft prompts or latent states, intermediate activations, and KV caches, including prefix caches reused by the current request. Its size and access pattern directly determine efficiency and generation quality. \emph{Compute-inactive memory} is information that does not currently participate in computation but can be stored, retrieved, updated, or forgotten: document collections, vector indexes, knowledge graphs, summary memories, user profiles, dialogue histories, and tool trajectories. It is cheaper to hold and suited to long-term maintenance, but influences generation only after being retrieved, compressed, and injected. The two are states rather than fixed categories: retrieved content becomes compute-active, and answers, plans, or reflections produced during inference return to the compute-inactive state when written back.

\paragraph{Intermediate states on the activation path.} KV offloading, persistent prefix caching, and retrieval-based attention give rise to information that lies between these two states. Such information has already been encoded into a form the model can consume directly (KV blocks, compressed latent states, or precomputed prefixes) and is kept in host memory, SSDs, or remote cache pools; it is not addressable by attention in the current request, yet it can be restored with low latency without re-retrieval or re-prefilling. Examples include offloaded KV states in InfiniGen~\cite{lee2024infinigen}, ShadowKV~\cite{sun2025shadowkv}, and RetrievalAttention~\cite{liu2024retrievalattention}, and compressed KV caches trained offline by Cartridges~\cite{eyuboglu2025cartridges}. A prefix cache retained for later requests is likewise in an intermediate state; it reaches this state either by being kept after a prefill (A $\to$ S) or by precomputation from text (I $\to$ S). We regard these as intermediate states on the transition path rather than as a third kind of memory. The criterion is the activation path: information that can return to computation through data movement and light correction is in an intermediate state, whereas information that additionally requires retrieval, compression, and re-encoding is compute-inactive. A transition between active and inactive memory may therefore pass through, and remain for some time in, an intermediate state with its own holding cost.

\subsection{State-Transition Operators}
\label{sec:operators}

Long-context compression methods can be described by a small set of operators, each specifying the direction in which information moves among compute-active (A), intermediate (S), and compute-inactive (I) states, possibly together with a change of representation such as summarization, structuring, or encoding into KV states (Table~\ref{tab:operators}).

\begin{table}[!tbp]
\centering
\scriptsize
\caption{State-transition operators. A: compute-active; S: intermediate state on the activation path; I: compute-inactive.}
\label{tab:operators}
\setlength{\tabcolsep}{3pt}
\renewcommand{\arraystretch}{1.15}
\begin{tabularx}{\textwidth}{@{}>{\raggedright\arraybackslash}p{2.0cm}>{\raggedright\arraybackslash}p{1.3cm}>{\raggedright\arraybackslash}p{3.1cm}>{\raggedright\arraybackslash}p{3.6cm}>{\raggedright\arraybackslash}X>{\raggedright\arraybackslash}p{1.2cm}@{}}
\toprule
\textbf{Operator} & \textbf{Transition} & \textbf{Typical representation change} & \textbf{Representative methods} & \textbf{Main cost / risk} & \textbf{Section} \\
\midrule
Write / externalize & A $\to$ I (copy) & Text $\to$ summaries, structured facts, or embeddings & MemoryBank~\cite{zhong2023memorybank}, Mem0~\cite{chhikara2025mem0}, A-MEM~\cite{xu2025amem} & Extraction and indexing cost; erroneous or sensitive content may be fixed & \ref{sec:writing}--\ref{sec:organization} \\
Offload & A $\to$ S & KV $\to$ KV (optionally quantized or low-rank) & InfiniGen~\cite{lee2024infinigen}, ShadowKV~\cite{sun2025shadowkv} & Host--device bandwidth and storage & \ref{sec:intermediate} \\
Recall & S $\to$ A & KV $\to$ KV & InfiniGen~\cite{lee2024infinigen}, RetrievalAttention~\cite{liu2024retrievalattention}, MagicPIG~\cite{chen2025magicpig} & Retrieval precision and transfer latency & \ref{sec:intermediate} \\
Retrieve-and-inject & I $\to$ A & Text / structure $\to$ text, latent, or KV & RAG~\cite{lewis2020rag}, LongLLMLingua~\cite{jiang-etal-2024-longllmlingua}, HippoRAG~\cite{gutierrez2024hipporag} & Retrieval, compression, and re-prefill cost & \ref{sec:retrieval}, \ref{sec:retrieved-to-active}--\ref{sec:injection-forms} \\
Materialize & I $\to$ S & Text $\to$ KV or latent & Prefix-cache precomputation, TurboRAG~\cite{lu2024turborag}, Cartridges~\cite{eyuboglu2025cartridges} & Offline compute and storage; tied to model version & \ref{sec:intermediate} \\
Evict & A $\to$ $\emptyset$ & --- & H2O~\cite{zhang2023h2o}, SnapKV~\cite{li2024snapkv}, StreamingLLM~\cite{xiao2024streamingllm} & Irrecoverable information loss & \ref{sec:kv} \\
Consolidate & within a state & Many items $\to$ summary, structure, or low-rank form & MiniCache~\cite{liu2024minicache} (A); SimpleMem~\cite{liu2026simplemem}, MemGAS~\cite{xu2025memgas} (I) & Loss of details and conditions through abstraction & \ref{sec:kv}, \ref{sec:maintenance} \\
Forget & I $\to$ $\emptyset$ or down-weight & --- & MemoryBank forgetting curve~\cite{zhong2023memorybank}, MemoryOS~\cite{kang2025memoryos} & Verifiable deletion and tracking of derived representations & \ref{sec:maintenance} \\
\bottomrule
\end{tabularx}
\end{table}

These operators are related to the atomic memory operations of \citet{du2025rethinking}, but are defined by changes of computational state rather than by memory function. This distinguishes \emph{retrieve-and-inject} (I $\to$ A, requiring re-prefilling) from \emph{recall} (S $\to$ A, mainly data movement), which both fetch memory but have very different cost structures. Compression, in turn, is not a separate operator but a representation change attached to the operators: eviction discards information within active memory, writing may summarize, retrieve-and-inject may filter and rewrite, and consolidation reorganizes representations within one state. Two further conventions apply. Writing copies rather than moves: written content usually remains in active memory for the rest of the request, so A $\to$ I creates a new representation in I without removing its source. Transformations that stay within compute-active memory, such as pruning, encoding into latent states, and architectural sharing or sparsification of KV states, change the representation or the accessed subset without changing state; Table~\ref{tab:methods} marks them as A $\to$ A.

\subsection{Information Lifecycle in Long-Context Systems}

Composing these operators yields an information lifecycle of writing, storage, retrieval, selection, compression, injection, computation, updating, and forgetting. A system first decides what to write and how to organize it as summaries, indexes, structured representations, or embeddings. For a given task, external memory is retrieved, but retrieved results are not yet effective context: they must be filtered, reranked, compressed, and formatted before entering the inference path. After inference, new facts, plans, tool feedback, user corrections, and reflections may be written back, while outdated, conflicting, low-value, or sensitive memories are updated, down-weighted, or forgotten. Long-context compression is therefore a continuous process rather than one-time input pruning, and different methods act at different stages of this lifecycle.

\subsection{A Budgeted Cost Model}
\label{sec:costmodel}

The aim of compression in this view is to improve information utilization under limited budgets, and the relevant objectives differ by state: for compute-active memory, prefill, KV memory, bandwidth, and throughput; for compute-inactive memory, preserving the retrievability and temporal validity of stored information; and for transitions, evidence traceability, recoverability, and long-term consistency. To make these objectives comparable, let $M$ be the set of memory units held at an inference step, and let each unit $m$ have a state $s(m) \in \{A, S, I\}$. The total cost is
\begin{equation}
C = \sum_{s(m)=A} c_A\,|m| + \sum_{s(m)=S} c_S\,|m| + \sum_{s(m)=I} c_I\,|m| + \sum_{\tau \in T} c_\tau ,
\label{eq:cost}
\end{equation}
where $|m|$ is the size of a unit (e.g., tokens or KV bytes); $c_A$, $c_S$, and $c_I$ are per-unit holding costs, typically with $c_A \gg c_S \gg c_I$, covering respectively prefill computation, KV memory, and decoding bandwidth; host-memory or SSD storage; and external storage and index maintenance; $T$ is the set of transitions performed for the current inference; and $c_\tau$ is the cost of each transition (retrieval, reranking, compression, data movement, recomputation, or write-back), which also depends on the representations before and after the transition. Long-context compression can then be written as
\begin{equation}
\max_{s(\cdot),\,T}\; U(y \mid M_A) - \lambda C \quad \text{s.t.} \quad \sum_{s(m)=A} |m| \le B_A,\quad \sum_{s(m)=S} |m| \le B_S ,
\label{eq:objective}
\end{equation}
where $U(y \mid M_A)$ is the utility of generating output $y$ from active memory $M_A$, $\lambda$ trades cost against utility, and $B_A$ and $B_S$ are the active and intermediate-state budgets.

The model exposes a common structure across method families. Prompt compression shrinks $M_A$ in text form; KV eviction removes units from $A$ at the price of recoverability; offloading and recall move units between $A$ and $S$, exchanging active budget for transfer cost; retrieval augmentation keeps information in $I$ and pays a retrieve-and-inject cost on demand; materialization pays an offline cost to move information from $I$ to $S$, lowering the cost of each later activation; and long-term memory management acts mainly on the $I$ term and its future activation cost. It also explains why compression ratio alone is misleading: a method may reduce the $c_A$ term while shifting cost into $c_\tau$ or offline maintenance.

\subsection{Analytical Dimensions for Method Comparison}

\begin{figure}[!tbp]
\centering
\scriptsize
\begin{forest}
  forked edges,
  for tree={
    grow'=0, draw, rounded corners=2pt, node options={align=left}, anchor=west, child anchor=west, parent anchor=east,
    inner xsep=3pt, inner ysep=1.5pt, l sep=7pt, s sep=2pt, fork sep=3pt, edge={draw=black!45},
    font=\scriptsize, line width=0.4pt
  },
  where level=0{text width=5.2cm, rotate=90, anchor=center, child anchor=north, node options={align=center}, font=\scriptsize\bfseries, fill=black!6}{},
  where level=1{text width=2.15cm, font=\scriptsize\bfseries}{},
  where level=2{text width=3.0cm}{},
  where level=3{text width=8.3cm, draw=black!30}{},
  where level=4{text width=5.55cm, draw=black!30}{},
  [{Memory-centric long-context compression}[{Compute-active memory (\S\ref{sec:active})}, fill=red!7, draw=red!45!black, for descendants={draw=red!35!black}
      [{Input and prompt compression (\S\ref{sec:prompt})}[{Selective Context; LLMLingua; LongLLMLingua}]]
      [{Compressed representations (\S\ref{sec:latent})}[{Gist Tokens; AutoCompressors; Activation Beacon; Infini-attention; LCLM}]]
      [{KV cache compression (\S\ref{sec:kv})}[{Heuristic eviction}, text width=2.6cm [{H2O; Scissorhands; StreamingLLM}]]
        [{Query-/head-aware}, text width=2.6cm [{SnapKV; FastKV; HeadKV; KV-Compress; Quest}]]
        [{Quantization and merging}, text width=2.6cm [{DiffKV; Lossless KV; MiniCache; xKV}]]
        [{Reusable KV}, text width=2.6cm [{KVzip; EchoKV}]]
        [{Reasoning traces}, text width=2.6cm [{R-KV; ThinKV}]]]
      [{Architecture-level selection (\S\ref{sec:arch})}[{MQA; GQA; MLA; NSA; MoBA; DeepSeek Sparse Attention; DeepSeek-V4; Kimi Linear; InfLLM; Mamba; Hymba}]]
      [{Serving-level management (\S\ref{sec:serving})}[{PagedAttention; vAttention; RadixAttention; Orca; Sarathi-Serve; DistServe; Mooncake; TokenLake; Jenga}]]]
    [{Compute-inactive memory (\S\ref{sec:inactive})}, fill=blue!6, draw=blue!45!black, for descendants={draw=blue!35!black}
      [{Memory writing (\S\ref{sec:writing})}[{MemoryBank; Generative Agents; ReAct trajectories; Reflexion; Voyager; Mem0; LightMem}]]
      [{Memory organization (\S\ref{sec:organization})}[{RETRO; MemGPT; MemoryOS; H-MEM; MemGAS; SimpleMem; A-MEM; Zep; MIRIX}]]
      [{Retrieval and activation readiness (\S\ref{sec:retrieval})}[{Dense retrieval (RAG); HippoRAG; HippoRAG 2; REMem}]]
      [{Update, consolidation and forgetting (\S\ref{sec:maintenance})}
        [{MemoryBank; MemoryOS; MemGAS; SimpleMem}]]
      [{Learned memory policies (\S\ref{sec:learned-policies})}[{Memory-R1}]]]
    [{State transitions (\S\ref{sec:transition})}, fill=violet!7, draw=violet!50!black, for descendants={draw=violet!35!black}
      [{Retrieve-and-inject (\S\ref{sec:retrieved-to-active}--\ref{sec:injection-forms})}[{LongLLMLingua; HippoRAG; TurboRAG; Block-Attention; CacheBlend}]]
      [{Offload and recall (\S\ref{sec:intermediate})}[{InfiniGen; ShadowKV; RetrievalAttention; MagicPIG; LongMem}]]
      [{Materialization (\S\ref{sec:intermediate})}[{Prefix caching; TurboRAG; Cartridges}]]
      [{Agent-side activation control (\S\ref{sec:agent-activation})}[{ACON; Context-Folding; MEM1; MemAgent}]]]
  ]
\end{forest}
\caption{Taxonomy of representative methods organized by memory state and state transition. Citations for all methods are given in the corresponding sections.}
\label{fig:taxonomy}
\end{figure}
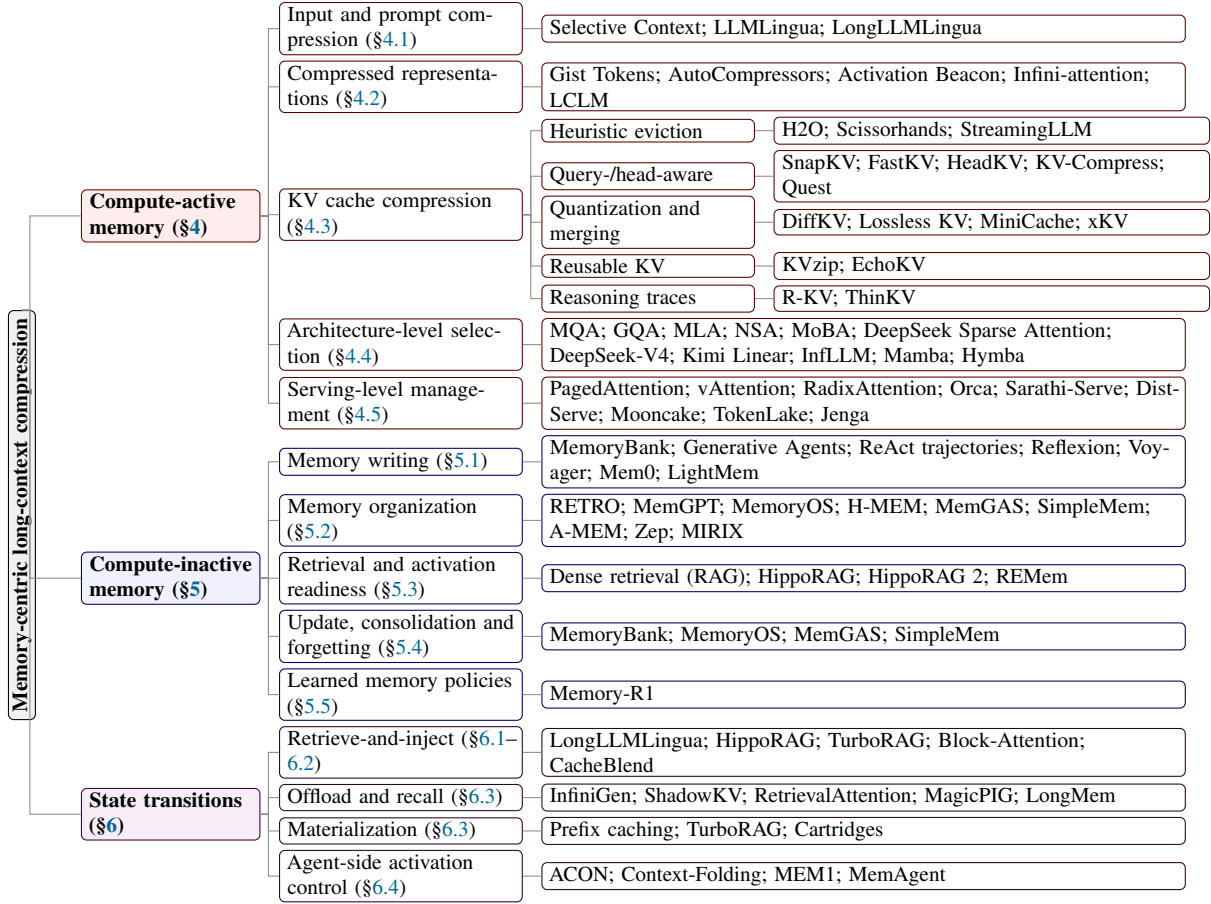

The following sections compare methods along shared dimensions: the memory state and operators involved; the compression target (text, structured fragments, continuous vectors, attention states, cache blocks, or indexes); the selection signal (query semantics, attention distributions, temporal state, task goals, retrieval scores, or heuristics); system requirements (access to internal states, training, compatibility with closed-model APIs, and cross-request reuse); and evaluation criteria (computational and storage cost, task utility, evidence fidelity, recoverability, and long-term consistency). Figure~\ref{fig:taxonomy} organizes representative methods by memory state and transition, and Table~\ref{tab:methods} codes a subset of them along these dimensions.

\begin{table}[!tbp]
\centering
\scriptsize
\caption{Representative methods coded along the analytical dimensions of Section~\ref{sec:framework}. Operators follow Table~\ref{tab:operators} (A: compute-active; S: intermediate; I: compute-inactive). Access: \emph{black-box} methods work with closed-model APIs; \emph{white-box} methods require model weights or internal states; \emph{system} methods operate in the serving runtime. Reuse: whether the compressed or stored state is reused across requests or sessions.}
\label{tab:methods}
\setlength{\tabcolsep}{3pt}
\renewcommand{\arraystretch}{1.08}
\begin{tabularx}{\textwidth}{@{}>{\raggedright\arraybackslash}p{2.6cm}>{\raggedright\arraybackslash}p{2.05cm}>{\raggedright\arraybackslash}p{1.05cm}>{\raggedright\arraybackslash}X>{\raggedright\arraybackslash}p{1.3cm}>{\raggedright\arraybackslash}p{1.35cm}>{\centering\arraybackslash}p{0.85cm}@{}}
\toprule
\textbf{Method} & \textbf{Operator} & \textbf{Repr.} & \textbf{Selection signal / mechanism} & \textbf{Training} & \textbf{Access} & \textbf{Reuse} \\
\midrule
\multicolumn{7}{@{}l}{\textit{Compute-active memory (Section~\ref{sec:active})}} \\
Selective Context & A$\to$A (prune) & Text & Self-information of lexical units & None & Black-box & -- \\
LLMLingua / LongLLMLingua & A$\to$A (prune) & Text & Small-LM perplexity under a budget; question-aware variant & None & Black-box & -- \\
Gist Tokens & A$\to$A (encode) & Latent & Learned gist tokens & Fine-tune & White-box & \cmark \\
AutoCompressors & A$\to$A (encode) & Latent & Segment-wise summary vectors & Fine-tune & White-box & \cmark \\
Activation Beacon & A$\to$A (encode) & KV & Learned activation compression & Fine-tune & White-box & -- \\
H2O / StreamingLLM & A$\to\emptyset$ (evict) & KV & Accumulated attention; attention sinks and recent window & None & White-box & -- \\
SnapKV / HeadKV & A$\to\emptyset$ (evict) & KV & Observation-window attention; head-level budgets & None & White-box & -- \\
MiniCache & Consolidate & KV & Cross-layer similarity merging & None & White-box & -- \\
KVzip & A$\to$A (compress) & KV & Query-agnostic context reconstruction & None & White-box & \cmark \\
R-KV & A$\to\emptyset$ (evict) & KV & Importance and redundancy in reasoning traces & None & White-box & -- \\
Quest & A$\to$A (select) & KV & Page-level attention upper bounds; full cache kept resident & None & White-box & -- \\
GQA / MLA & A$\to$A (arch.) & KV & Shared or low-rank latent KV heads & Pre-train or uptrain & White-box & -- \\
NSA / MoBA & A$\to$A (arch.) & KV & Learned block compression, selection, routing & Pre-train & White-box & -- \\
Kimi Linear & A$\to$A (arch.) & KV & Hybrid linear and full attention & Pre-train & White-box & -- \\
PagedAttention & -- & KV & Block-level KV memory allocation & None & System & -- \\
RadixAttention & -- & KV & Prefix matching in a radix tree & None & System & \cmark \\
Mooncake & A$\leftrightarrow$S & KV & KV-centric disaggregated scheduling & None & System & \cmark \\
\midrule
\multicolumn{7}{@{}l}{\textit{Compute-inactive memory (Section~\ref{sec:inactive})}} \\
MemoryBank & A$\to$I, forget & Text & Ebbinghaus-style forgetting curve & None & Black-box & \cmark \\
Generative Agents & A$\to$I, I$\to$A & Text & Recency, importance, and relevance & None & Black-box & \cmark \\
MemGPT & A$\leftrightarrow$I & Text & LLM-directed paging via function calls & None & Black-box & \cmark \\
HippoRAG & I$\to$A & Graph & Knowledge graph and personalized PageRank & None & Black-box & \cmark \\
Reflexion & A$\to$I & Text & Verbal self-reflection on feedback & None & Black-box & \cmark \\
Mem0 & A$\to$I, consolidate & Text / graph & LLM-decided add, update, delete & None & Black-box & \cmark \\
A-MEM & A$\to$I, consolidate & Struct. & Linked atomic notes & None & Black-box & \cmark \\
Zep & A$\to$I & Graph & Temporal knowledge graph & None & Black-box & \cmark \\
Memory-R1 & A$\to$I, consolidate & Text & RL-learned memory operations & RL & White-box & \cmark \\
\midrule
\multicolumn{7}{@{}l}{\textit{State transitions (Section~\ref{sec:transition})}} \\
TurboRAG / Block-Attention & I$\to$S$\to$A & KV & Precomputed chunk KV; block-wise independent encoding & Fine-tune & White-box & \cmark \\
CacheBlend & I$\to$S$\to$A & KV & Selective recomputation of deviating tokens & None & White-box & \cmark \\
Cartridges & I$\to$S & KV & Self-study distillation of a corpus & Per-corpus & White-box & \cmark \\
InfiniGen & A$\leftrightarrow$S & KV & Next-layer KV prefetch predicted from the previous layer & None & White-box & -- \\
RetrievalAttention / ShadowKV & A$\leftrightarrow$S & KV & ANN index over keys; low-rank keys with offloaded values & None & White-box & -- \\
ACON & A$\to$A (summarize) & Text & Guidelines optimized from failure cases; optional distilled compressor & None (prompt opt.) & Black-box & -- \\
Context-Folding & A$\to$A (fold) & Text & RL-learned branching and folding & RL & White-box & -- \\
MEM1 / MemAgent & A$\to$A (overwrite) & Text & RL-learned bounded internal memory & RL & White-box & -- \\
\bottomrule
\end{tabularx}
\end{table}

\section{Compressing Compute-Active Memory}
\label{sec:active}

Compute-active memory compression reduces the cost of information that is already on the inference path, and therefore acts directly on prefill overhead, decoding latency, GPU memory, and throughput. We organize methods by where compression occurs: text before model input, compact latent representations, the KV cache during decoding, the model architecture, and the serving system. For each family we consider which cost is reduced, what information is preserved, and what the method requires of model access, training, and deployment. KV states that leave the attention path through offloading are discussed with other state transitions in Section~\ref{sec:intermediate}.

\subsection{Input and Prompt Compression}
\label{sec:prompt}

Input and prompt compression filters or rewrites context before it enters the model, acting as admission control for compute-active memory. Inputs that concatenate documents, interaction histories, retrieved passages, and instructions often contain repeated background, templated text, and weakly relevant passages that raise prefill cost and dilute useful information. Methods differ in how they estimate information value: redundancy-based methods such as Selective Context~\cite{li-etal-2023-compressing} remove locally repetitive or low-information tokens, sentences, or passages; LLMLingua~\cite{jiang-etal-2023-llmlingua} formulates compression as coarse-to-fine selection under a token budget; and LongLLMLingua~\cite{jiang-etal-2024-longllmlingua} conditions compression on the query to retain relevant evidence in long contexts.

Because the output remains natural language, these methods are inspectable and require no change to model internals. Readability does not by itself ensure evidence traceability, however: token-level pruning may produce ungrammatical text detached from its source, so applications such as RAG and knowledge-base question answering also need to retain the original positions, sources, and timestamps of the selected content. Their limitation is that compression precedes reasoning: content that appears unimportant locally may be needed for multi-hop reasoning, long-range coreference, code dependencies, or cross-turn consistency. They work best when task goals are clear, redundancy is high, and evidence can be located from the query, and otherwise need to be combined with reranking, structured evidence preservation, or downstream verification.

\subsection{Compressed Representations: Gist Tokens, Latent States, and Summary Memory}
\label{sec:latent}

Compressed-representation methods encode long contexts into denser computable states, such as special tokens, soft prompts, summary vectors, or compressed activations, which still participate in inference but occupy far less of the context budget. Three directions can be distinguished. Gist Tokens~\cite{mu2023learning} train a model to store prompt information in a few learned tokens that later generation attends to; AutoCompressors~\cite{chevalier-etal-2023-adapting} process long texts segment by segment and pass summary vectors forward; and Activation Beacon~\cite{zhang2025long} and Infini-attention~\cite{munkhdalai2024leave} compress Transformer activations or combine local attention with a bounded long-term compressed memory. More recent work scales encoder--decoder context compression to large-scale training, improving the trade-off among task performance, compression speed, and peak memory~\cite{li2026lclm}.

These representations increase information density and avoid repeated encoding when the same background is accessed many times. They are, however, difficult to interpret: it is hard to determine which facts or constraints a latent state preserves, or where information was lost when generation fails. They are also tied to specific architectures and training objectives, which limits transfer across models and to closed APIs. They are best viewed as high-density intermediate representations to be combined with task-aware compression or traceable retrieval.

\subsection{KV Cache Compression}
\label{sec:kv}

During autoregressive decoding, the keys and values of past tokens are cached to avoid recomputing the prefix, and this cache grows with context length, depth, number of heads, and batch size until it dominates GPU memory and bandwidth. The KV cache is thus a high-cost form of short-term compute-active memory, and KV compression seeks to retain the states most useful for future generation within a fixed budget. Methods differ mainly in how they estimate future utility: by recency, accumulated attention, or attention sinks~\cite{zhang2023h2o,liu2023scissorhands,xiao2024streamingllm}; by query-, head-, or layer-aware importance~\cite{li2024snapkv,jo2026fastkv,fu2025headkv,rehg2024kvcompress}; or by reducing representation cost through quantization, merging, low-rank reconstruction, or reuse across queries~\cite{zhang2025diffkv,yang2024losslesskv,liu2024minicache,chang2026xkv,kim2025kvzip,wang2026echokv}. Not all methods discard states: Quest~\cite{tang2024quest} keeps the full cache resident and loads only the pages with the highest attention upper bounds for the current query, reducing reads and attention computation rather than memory capacity. Systematic analyses further show that compression can degrade capabilities unevenly, even when aggregate scores are preserved~\cite{liu2026semanticintegrity}. These methods exploit internal signals unavailable to text-level compression, but require access to model states and inference-system support, and their decisions rest on imperfect predictions: states that received little attention so far may become critical later.

A fast-growing branch targets reasoning models, for which long chains of thought shift the main source of KV growth from the prompt to the model's own generations, and redundancy arises from repeated verification and self-correction. R-KV~\cite{cai2025rkv} compresses reasoning-trace KV states by jointly considering attention importance and redundancy, and ThinKV~\cite{ramachandran2026thinkv} allocates cache budgets and quantization precision according to the type of thought segment. These methods show that compute-active memory grows not only with external input but also with the reasoning process itself.

\subsection{Architecture-level Selective Computation}
\label{sec:arch}

Architecture-level methods reduce the cost of long contexts by changing how the model represents and accesses history, rather than by deleting input or post-processing the cache. Multi-Query and Grouped-Query Attention~\cite{shazeer2019mqa,ainslie-etal-2023-gqa} share keys and values across query heads, and Multi-head Latent Attention~\cite{deepseekai2024deepseekv2,deepseekai2024deepseekv3} compresses KV states into latent vectors. Sparse attention has since become native and trainable: NSA~\cite{yuan2025nsa} combines compression, selection, and sliding-window branches in a hardware-aligned design, MoBA~\cite{lu2025moba} routes queries to attention blocks in a mixture-of-experts fashion, and DeepSeek-V3.2~\cite{deepseekai2025v32} introduces sparse attention driven by a lightweight indexer; DeepSeek-V4~\cite{deepseekai2026deepseekv4} further integrates compressed attention, shared KV, and local windows into the attention mechanism. Other designs reduce reliance on full attention altogether: InfLLM~\cite{xiao2024infllm} retrieves from an efficient context memory, state-space and hybrid models such as Mamba~\cite{gu2023mamba} and Hymba~\cite{dong2024hymba} maintain recurrent states, and Kimi Linear~\cite{kimi2025linear} pairs a few full-attention layers with linear-attention layers to shrink the KV cache substantially.

These methods build the selection or compaction of compute-active memory into the attention mechanism itself. Most require pre-training or continued training, although GQA can also be obtained by uptraining an existing multi-head checkpoint~\cite{ainslie-etal-2023-gqa}, and InfLLM is training-free and operates on a pre-trained model at inference time~\cite{xiao2024infllm}. We include context-extension techniques only when they change the access pattern, state representation, or cost structure of inference. Their main limitation is cost of adoption: they require changes to the architecture or attention kernels and to cache management, usually together with training, and cannot be applied to closed-model interfaces.

\subsection{Serving-level Cache and Prefix Management}
\label{sec:serving}

Serving-level methods do not change semantic content or internal representations; they determine how computed states are placed, reused, scheduled, and released across GPU memory, cache blocks, batches, and requests. vLLM/PagedAttention~\cite{kwon2023pagedattention} manages KV caches in pages to reduce fragmentation and enable sharing, and vAttention~\cite{prabhu2024vattention} achieves similar benefits by decoupling virtual and physical memory. Prefix caching stores the KV states of stable contexts for reuse, reducing time-to-first-token, and SGLang/RadixAttention~\cite{zheng2024sglang} automatically identifies and reuses shared prefixes across multi-call programs, branching, and multi-turn interaction. Because prefill is compute-bound and decoding is bandwidth-bound, systems also schedule across phases: Orca introduces iteration-level scheduling~\cite{yu2022orca}, Sarathi-Serve uses chunked prefill~\cite{agrawal2024sarathi}, DistServe disaggregates prefill and decoding~\cite{zhong2024distserve}, and Mooncake builds a disaggregated KV cache across CPU, DRAM, and SSD resources~\cite{qin2025mooncake}. TokenLake extends prefix caching to a segment-level cache pool, and Jenga manages memory under heterogeneous cache layouts~\cite{wu2025tokenlake,zhang2025jenga}.

Serving-level management is what converts the local savings of other methods into end-to-end latency, throughput, and memory gains. Its benefits depend strongly on frameworks, hardware, workloads, and cache hit rates, so evaluation should report throughput, memory utilization, hit rates, and tail latency rather than single-request speed alone.

\section{Compute-Inactive Memory Compression}
\label{sec:inactive}

Not all valuable information needs to remain in the attention window. Source documents, dialogue histories, user preferences, tool outputs, task states, and reflections are often stored externally and retrieved when needed. Such compute-inactive memory incurs no prefill or KV cost at the current step, but determines whether a system can maintain continuity and reuse experience over long-term operation. Compressing it therefore means turning large histories into memory structures that remain retrievable, updatable, and reactivatable, not merely reducing storage. This section follows the external-memory lifecycle: writing, organization, retrieval, maintenance, and learned policies that govern these steps; how retrieved memory is compressed and injected into the inference path is discussed in Section~\ref{sec:transition}.

\subsection{Memory Writing}
\label{sec:writing}

Writing determines which parts of the current interaction, task execution, or tool feedback become long-term memory units, and is the first selection boundary of an external-memory system. Writing complete histories reduces omission but enlarges the store, adds retrieval noise, and amplifies privacy risks and the accumulation of stale or erroneous content. Written objects can be grouped into five types: factual memory of user preferences and stable states~\cite{zhong2023memorybank,kang2025memoryos}; episodic memory of temporally grounded experiences~\cite{park2023generativeagents,shu2026remem}; trajectory memory of reasoning steps, tool calls, and feedback~\cite{yao2023react}; reflective memory that abstracts experiences or failures into lessons~\cite{shinn2023reflexion}; and skill memory that consolidates successful experience into callable procedures~\cite{wang2023voyager}. Recent systems make writing an explicit decision: Mem0~\cite{chhikara2025mem0} extracts candidate facts after each interaction and decides whether to add, update, delete, or keep them, and LightMem~\cite{fang2025lightmem} pre-compresses and topically segments input before writing and defers consolidation to an offline stage. Writing is a selection made under uncertainty about future tasks: overly conservative writing weakens continuity, whereas overly aggressive writing lowers the signal-to-noise ratio of the store on which all later stages depend.

\subsection{Memory Organization}
\label{sec:organization}

Organization determines how written memories are indexed, related, and later updated. Appending memories to a linear log preserves history but lets redundancy, staleness, and retrieval noise grow with interaction length. Existing systems use serialized memory streams that preserve temporal order~\cite{park2023generativeagents}, vector indexes for semantic retrieval~\cite{lewis2020rag,borgeaud2022retro}, external stores of independently accessible units~\cite{zhong2023memorybank,packer2023memgpt}, graph-structured memory of entities and relations~\cite{gutierrez2024hipporag,gutierrez2025fromrag}, hierarchical and multi-granularity memory across time scales~\cite{kang2025memoryos,sun2025hmem,xu2025memgas}, structured semantic compression of memory units~\cite{liu2026simplemem}, and stores of reflections and skills~\cite{shinn2023reflexion,wang2023voyager}. Recent systems refine these forms: A-MEM~\cite{xu2025amem} organizes memories as linked atomic notes that evolve with new information, Zep~\cite{rasmussen2025zep} uses a temporal knowledge graph with validity intervals for facts, and MIRIX~\cite{wang2025mirix} separates core, episodic, semantic, procedural, and resource memory managed by multiple agents.

Organization is itself a form of compression: without necessarily reducing the number of entries, structuring, indexing, and hierarchization reduce the noise faced by later retrieval and injection. The cost is maintenance: richer structures make updating, deletion, and conflict correction harder, and memory units that are too coarse import irrelevant content while units that are too fine break contextual integrity.

\subsection{Memory Retrieval and Activation Readiness}
\label{sec:retrieval}

Retrieval decides which memories are relevant to the current query, task state, or agent goal. Vector retrieval relies on semantic similarity~\cite{lewis2020rag,borgeaud2022retro}; interactive agents additionally weight recency and importance~\cite{park2023generativeagents,zhong2023memorybank}; memory controllers schedule access across memory tiers~\cite{packer2023memgpt,kang2025memoryos}; graph-based retrieval follows entity and relation paths for multi-hop and temporal questions~\cite{gutierrez2024hipporag,gutierrez2025fromrag}; episodic retrieval recalls specific past experiences~\cite{shu2026remem,chhikara2025mem0}; and agent memory may be triggered by task goals, tool states, or prior failures~\cite{shinn2023reflexion,wang2023voyager}. Retrieval compresses the candidate space so that the full history need not be exposed in context, and its quality bounds every later step: retrieving too little causes forgetting and inconsistency, whereas retrieving too much reintroduces noise and increases the cost of later compression and injection. We refer to this stage as \emph{activation readiness} because retrieved memories remain candidates; they become compute-active only after the selection, compression, and injection steps discussed in Section~\ref{sec:transition}.

\subsection{Memory Update, Consolidation and Forgetting}
\label{sec:maintenance}

Long-term memory does not remain correct indefinitely: preferences change, task states evolve, facts become stale, and sources conflict. \emph{Updating} reconciles new information with stored memory, reinforcing consistent entries and, for conflicts, replacing, revising, versioning, or deferring until confirmation; without it, stale information is repeatedly reactivated. \emph{Consolidation} merges repeated or similar memories into summaries, topics, structured facts, or higher-level reflections, increasing density at the risk of losing details, temporal order, or conditions that later tasks need. \emph{Forgetting} removes, down-weights, expires, hides, or archives memories that are low-value, stale, erroneous, or sensitive, serving both noise control and privacy and safety governance. Together these operations determine whether compute-inactive memory keeps a high signal-to-noise ratio over time; a system that only writes and retrieves gradually degrades into a store of historical noise.

\subsection{Learned Memory-Management Policies}
\label{sec:learned-policies}

The mechanisms above mostly rely on hand-designed rules or prompts. A recent trend is to learn memory-management policies with reinforcement learning, so that what to store, revise, and use is optimized by task reward. Memory-R1~\cite{yan2025memoryr1} trains a memory manager to add, update, delete, or skip entries in an external memory bank and an answer agent to filter retrieved memories, turning the write, consolidate, and forget operators into learned policies that maximize utility under a budget, as in Eq.~\eqref{eq:objective}. Related methods learn to maintain a bounded memory inside the context window; because that memory is compute-active by our criterion, we discuss them with agent-side activation control (Section~\ref{sec:agent-activation}). Because rewards usually come only from final answers, however, such learned policies do not directly constrain the faithfulness, traceability, or privacy of memory, and may exploit benchmark regularities; incorporating cost and reliability terms into the reward remains open.

\section{State Transitions: Retrieval, Injection, and Intermediate States}
\label{sec:transition}

This section examines how information moves between memory states, which is where the memory-centric view departs most from existing reviews. Retrieved memory is not equivalent to information the model has used: it affects generation only after selection, compression, formatting, and injection into the attention path. Injecting every candidate reintroduces the cost and noise of long contexts, whereas aggressive filtering can remove evidence, temporal relations, or task state. We first discuss the transition from retrieved memory to active context and the forms of injection, then intermediate states on the activation path, agent-side activation control, and the feedback loop created by write-back.

\subsection{From Retrieved Memory to Active Context}
\label{sec:retrieved-to-active}

Retrieval is often presented as a way to avoid placing all history in the input, but it is not a cost-free extension of context: as soon as retrieved content is concatenated into the prompt, it consumes prefill computation, context length, and subsequent cache budget. If candidates are injected wholesale, redundant, weakly relevant, or stale content re-enters the inference path and erodes the gains of compression. The transition from candidates to active context therefore requires three judgments: which candidates are relevant to the current task, which can be summarized, structured, or merged, and which must be kept as original evidence. The answer depends on the task: long-document question answering needs verifiable evidence, personalized dialogue needs consistent user state, code tasks need structural dependencies, and agent tasks rely on tool results and action histories. Retrieval-conditioned compression and injection is thus an activation-control process that connects the external memory of Section~\ref{sec:inactive} with the active memory of Section~\ref{sec:active}.

\subsection{Compression and Injection Forms}
\label{sec:injection-forms}

Post-retrieval compression operates at several granularities: filtering removes weakly relevant candidates, reranking orders evidence, summarization merges candidates into shorter text, structured extraction produces entities, attributes, temporal or causal relations, and task states, graph-path compression keeps the key nodes and edges of multi-hop relations, and more aggressive methods encode retrieved content as special tokens, latent vectors, or cache states. Each choice carries its own risk: summaries are inspectable but may omit detail, structured facts are compact but depend on extraction quality, and latent representations compress most but are least interpretable.

The injection form determines the computational state in which external memory participates. Concatenating passages is simple and interpretable but costly; summaries are shorter but lossy; structured facts and graph paths suit relational, temporal, and multi-hop reasoning; and in more integrated systems, external memory enters as soft prompts, latent vectors, prefix states, or KV caches. At the KV level, TurboRAG~\cite{lu2024turborag} precomputes KV caches for document chunks offline and concatenates them online to skip prefilling of retrieved content; Block-Attention~\cite{sun2025blockattention} encodes retrieved blocks independently, lets only the final query block attend to all blocks, and fine-tunes the model to this pattern; and CacheBlend~\cite{yao2025cacheblend} restores cross-document attention lost by naive concatenation by recomputing KV states for a small set of tokens with the largest deviation. No injection form is uniformly superior: traceable evidence favors text or structured facts, extreme compression favors latent representations, and multi-turn reuse favors prefix or cache states, so injection must be chosen jointly by cost, utility, and verifiability.

\subsection{Intermediate States on the Activation Path: Offloading and Materialization}
\label{sec:intermediate}

Retrieve-and-inject requires retrieval, compression, and re-encoding before external memory can be used. Intermediate states avoid the last two steps: the information already has a form the model can use and returns to computation through data movement and light correction. Such states arise in two directions, by offloading active KV states from the attention path (offload and recall in Table~\ref{tab:operators}) or by materializing external memory in advance into loadable forms (materialize).

\paragraph{Offloading and recall.} When historical KV states exceed the GPU budget, a system can move them to host memory or lower storage tiers instead of evicting them, and recall a small relevant subset during decoding. The key problem is to decide what to recall without touching all states. InfiniGen~\cite{lee2024infinigen} predicts important tokens of the next layer from the previous layer's input and partial query--key projections and prefetches their KV states; RetrievalAttention~\cite{liu2024retrievalattention} builds attention-aware vector indexes that account for the distribution mismatch between queries and keys; MagicPIG~\cite{chen2025magicpig} estimates attention by locality-sensitive-hashing sampling rather than top-$k$ selection; and ShadowKV~\cite{sun2025shadowkv} keeps low-rank keys on the GPU, offloads values to the CPU, and reconstructs selected keys during decoding. LongMem~\cite{wang2023longmem} applies the same idea beyond a single input, caching attention key--value pairs of past inputs in a memory bank that a trained side network retrieves from. Quest~\cite{tang2024quest} uses query-aware page selection as well but keeps the full cache in accelerator memory, so it changes which states are read rather than where they reside (Section~\ref{sec:kv}). These methods are mechanistically isomorphic to external-memory retrieval---both index information beyond the active budget, select candidates for the current query, and reactivate them at a cost---but KV recall requires no re-prefilling. KV-level retrieval-based attention and text-level retrieval augmentation can thus be viewed as instances of one activation problem over different representations. They remain sensitive to host--device bandwidth, index construction, and recall precision, and mostly operate within a single request.

\paragraph{Materialization.} In the other direction, external memory can be moved in advance from the compute-inactive state into an intermediate state, so that each later activation pays only for loading and correction. Cross-request prefix caching (Section~\ref{sec:serving}) is the simplest case, and the KV-level injection methods of Section~\ref{sec:injection-forms} follow the same idea, facing the mismatch between independently encoded chunks and joint reasoning. Cartridges~\cite{eyuboglu2025cartridges} take a different route: the model generates synthetic conversations about a corpus and distills it into a much smaller trainable KV cache, showing that materialization can be a lossy, learned compression targeted at the expected query distribution rather than lossless storage of raw KV states. Materialized states raise their own problems: they are bound to a specific model version and must be rebuilt after updates, they occupy far more storage than raw text, and they are hard to inspect or edit, so invalidating the derived states of a deleted or corrected document becomes a system-level instance of the governance problems in Section~\ref{sec:challenges}.

\subsection{Agent-Side Activation Control}
\label{sec:agent-activation}

In long-horizon agent tasks, activation control also applies to the agent's own trajectory: as tool calls, observations, and intermediate reasoning accumulate, the agent must decide which history stays in active context and which is summarized, folded, or externalized. ACON~\cite{kang2025acon} refines compression guidelines from trajectories in which the full context succeeds but the compressed context fails, and distills the resulting compressor into smaller models; Context-Folding~\cite{sun2025contextfolding} lets an agent branch into sub-tasks and fold each completed sub-trajectory into a summary, learning the folding behavior with reinforcement learning; MEM1~\cite{zhou2025mem1} lets an agent merge its previous internal state with each new observation into a compact state, keeping memory nearly constant in long-horizon tasks; and MemAgent~\cite{yu2025memagent} reads long documents segment by segment and learns to overwrite a bounded in-context memory, allowing a limited-context model to handle very long inputs. Here the decision-maker shifts from external retrieval modules to the agent policy, and when and what to compress becomes part of the action space. A dedicated survey~\cite{ctxcomp2026agents} reviews these methods in detail; from the lifecycle perspective, the open questions are whether folded trajectories should be written back to long-term memory, in what form, and whether information lost in folding can later be recovered.

\subsection{Feedback and Failure Modes}

Transitions form a loop: injected memory shapes current inference, and model outputs, tool feedback, user corrections, and reflections are written back as new memory. The loop allows experience to accumulate, but it also allows errors to persist. Retrieval may return stale or weakly relevant memories, compression may delete critical evidence, injection may use a format the model cannot exploit, inference may over-rely on injected content, and write-back may store erroneous outputs or low-quality reflections, which are then retrieved again in future tasks. The loop also has a security dimension: memory-poisoning attacks implant malicious entries into memory or knowledge bases that are later retrieved under specific triggers, as demonstrated by AgentPoison~\cite{chen2024agentpoison}, and a recent survey on long-term memory security~\cite{lin2026memsecurity} organizes such threats across the write, store, retrieve, execute, share, and forget phases and notes that compression can amplify poisoned content. Since every transition is a potential point of error or attack, validation must cover writing, materialization, injection, and write-back.

\section{Evaluation, Benchmarks, and Metrics}
\label{sec:evaluation}

Evaluating long-context compression requires more than measuring how much information is removed. A method that shortens the input, shrinks the cache, or retrieves less is not effective if it damages evidence, reasoning chains, user state, or long-term consistency. Because costs arise at different stages---prefill for input compression, decoding memory for KV compression, recall and summarization for retrieval-conditioned compression, and writing, indexing, and maintenance for long-term memory---evaluation should report cost--quality curves across context lengths and budgets, so that genuine improvements can be distinguished from costs shifted into external storage or background maintenance.

\subsection{Cost, Utility, and Long-term Reliability}

\emph{Cost} should be reported beyond compression ratio: input tokens, cache size, peak GPU memory, prefill latency, time-to-first-token, decoding latency, retrieval and compression latency, end-to-end latency, and throughput. Following the cost model in Section~\ref{sec:costmodel}, we recommend reporting separately the active holding cost (prefill tokens and peak KV memory), the intermediate-state holding cost (offloaded or materialized storage), the transition cost (retrieval, transfer, compression, and recomputation), and offline maintenance cost, together with the number of requests over which each cost is amortized; vectorization, graph construction, consolidation, and prefix precomputation can otherwise move inference cost into unreported background processes.

\emph{Task utility} should extend beyond final accuracy to evidence retention, faithfulness, reasoning consistency, hallucination rate, and task success. Question answering should be grounded in the correct evidence, structured and graph memory should preserve relation paths and state constraints, and agent tasks should additionally be assessed on planning, tool use, failure recovery, and skill reuse.

\emph{Long-term reliability} distinguishes memory-centric compression from generic input or cache compression. Evaluation should test whether a system writes new memories correctly, retrieves old ones at the right time, updates stale states, consolidates repeated memories, and down-weights or forgets low-value, erroneous, sensitive, or retracted information. A method that optimizes single answers while accumulating erroneous reflections, stale preferences, or conflicting states is not a reliable long-term solution.

\subsection{Benchmark Gaps and Evaluation Protocols}

Existing benchmarks cover important but partial capabilities (Table~\ref{tab:benchmarks}). Comprehensive long-context benchmarks test understanding, question answering, summarization, code, and reasoning over long documents; synthetic retrieval tasks test localization at extreme lengths; and long-term memory benchmarks examine cross-session recall, temporal reasoning, knowledge updates, and implicit conflicts. Recent benchmarks address further gaps. InfiniteBench~\cite{zhang2024infinitebench} extends evaluation beyond 100K tokens; HELMET~\cite{yen2025helmet} shows that synthetic tasks correlate weakly with downstream performance and adds application-oriented tasks; NoLiMa~\cite{modarressi2025nolima} removes literal overlap between questions and evidence and reveals sharp degradation when semantic association is required; BABILong~\cite{kuratov2024babilong} embeds multi-fact reasoning in long distractor text; SCBench~\cite{li2025scbench} evaluates shared-context, multi-request settings across the generation, compression, retrieval, and loading of KV caches; MemoryAgentBench~\cite{hu2026memoryagentbench} evaluates retrieval, test-time learning, long-range understanding, and conflict resolution in incremental interaction; and STALE~\cite{chao2026stale} tests whether agents recognize memories that are no longer valid.

These benchmarks remain dominated by static inputs or isolated capabilities. Success on needle-style retrieval does not imply multi-evidence reasoning, accuracy on long-document question answering does not imply maintenance of user state, and recalling historical facts does not imply correct updating or forgetting. Evaluation protocols should therefore separate context comprehension, evidence localization, memory retrieval, state updating, forgetting, and agent execution, and should vary context length, evidence position, compression, cache, retrieval, and injection budgets, memory scale, and interaction rounds jointly while reporting both stage-level and end-to-end cost.

\begin{table}[!tbp]
\centering
\scriptsize
\caption{Coverage of representative benchmarks across lifecycle capabilities. \oldCIRCLE: primary focus; \oldLEFTcircle: partially covered; \oldCircle: not covered. No benchmark jointly covers KV-level compression, memory updating, and systematic cost reporting.}
\label{tab:benchmarks}
\setlength{\tabcolsep}{3pt}
\renewcommand{\arraystretch}{1.12}
\begin{tabularx}{\textwidth}{@{}>{\raggedright\arraybackslash}p{3.1cm}>{\raggedright\arraybackslash}X*{7}{>{\centering\arraybackslash}p{1.05cm}}@{}}
\toprule
\textbf{Benchmark} & \textbf{Setting} & \textbf{Localiza\-tion} & \textbf{Multi-evid.\ reasoning} & \textbf{Cross-session memory} & \textbf{Update / conflict} & \textbf{Compres\-sion-aware} & \textbf{Agent / tool} & \textbf{Cost report} \\
\midrule
NIAH / Passkey~\cite{yu2025sequentialniah,munkhdalai2024leave} & Synthetic long sequences & \CIRCLE & \Circle & \Circle & \Circle & \Circle & \Circle & \Circle \\
RULER~\cite{hsieh2024ruler} & Controlled lengths & \CIRCLE & \LEFTcircle & \Circle & \Circle & \Circle & \Circle & \Circle \\
NoLiMa~\cite{modarressi2025nolima} & Non-literal matching & \CIRCLE & \LEFTcircle & \Circle & \Circle & \Circle & \Circle & \Circle \\
BABILong~\cite{kuratov2024babilong} & Facts in distractor text & \LEFTcircle & \CIRCLE & \Circle & \Circle & \Circle & \Circle & \Circle \\
LongBench / v2~\cite{bai2024longbench,bai2024longbenchv2} & Natural long documents & \LEFTcircle & \CIRCLE & \Circle & \Circle & \Circle & \Circle & \Circle \\
InfiniteBench~\cite{zhang2024infinitebench} & Beyond 100K tokens & \LEFTcircle & \CIRCLE & \Circle & \Circle & \Circle & \Circle & \Circle \\
HELMET~\cite{yen2025helmet} & Application-oriented tasks & \LEFTcircle & \CIRCLE & \Circle & \Circle & \Circle & \Circle & \Circle \\
SCBench~\cite{li2025scbench} & Shared context, multi-request & \LEFTcircle & \LEFTcircle & \LEFTcircle & \Circle & \CIRCLE & \Circle & \LEFTcircle \\
LoCoMo~\cite{maharana2024locomo} & Multi-session dialogue & \LEFTcircle & \LEFTcircle & \CIRCLE & \LEFTcircle & \Circle & \Circle & \Circle \\
LongMemEval~\cite{wu2024longmemeval} & Multi-session chat history & \LEFTcircle & \LEFTcircle & \CIRCLE & \CIRCLE & \Circle & \Circle & \Circle \\
MemoryAgentBench~\cite{hu2026memoryagentbench} & Incremental interaction & \LEFTcircle & \LEFTcircle & \CIRCLE & \CIRCLE & \Circle & \LEFTcircle & \Circle \\
STALE~\cite{chao2026stale} & Evolving facts and states & \Circle & \LEFTcircle & \LEFTcircle & \CIRCLE & \Circle & \LEFTcircle & \Circle \\
Code / agent benchmarks~\cite{liu2024repobench,jimenez2023swebench,zhou2023webarena} & Repositories, web, tools & \LEFTcircle & \CIRCLE & \Circle & \Circle & \Circle & \CIRCLE & \Circle \\
\bottomrule
\end{tabularx}
\end{table}

\section{Open Challenges and Future Directions}
\label{sec:challenges}

\subsection{Learned, Cross-Tier Budget Allocation}

Most methods rely on fixed policies, such as a fixed compression ratio, number of retrieved items, cache budget, or summary granularity, although memory requirements vary with the task: factual question answering needs verifiable evidence, summarization needs coverage, code tasks need interfaces and dependencies, long-term dialogue needs temporally valid user state, and agents need goals, tool results, and reusable experience. Components are also optimized in isolation even though they interact: excessive retrieval enlarges the current context, aggressive context compression degrades both generation and subsequent write-back, and KV eviction determines whether activated information remains usable. Allocation should therefore be decided jointly across input compression, KV retention and offloading, retrieval, post-retrieval compression, and injection form, according to task goals, model uncertainty, and the available budget.

\paragraph{Research question.} Can a single allocation policy be trained with the cost model of Section~\ref{sec:costmodel} as its objective, jointly choosing the prompt compression ratio, the fraction of KV states retained or offloaded, the number of retrieved items, and the injection form within a request? The learned policies in Sections~\ref{sec:learned-policies} and~\ref{sec:agent-activation} show that memory decisions can be learned, but their rewards include no explicit cost terms and do not cover KV- or serving-level decisions.

\subsection{Trustworthy Write-back, Security, and Governable Forgetting}

Errors in long-term memory are dangerous because they persist. Stale or irrelevant retrievals, compression that drops conditions, misattributed reflections, and misread tool results can be written back and reactivated in later tasks, turning a local error into a lasting bias; poisoning attacks exploit the same path deliberately. Memories should therefore not be retained merely because they appear in dialogue, model output, or tool logs; write-back requires evidence checking, conflict detection, provenance tracking, and versioning, and must distinguish stable facts from temporary states and uncertain inferences. Long-term memory also stores preferences, histories, and potentially sensitive personal information, so users need to know what is remembered, why, and when it is used, and to correct or delete it. Because memory is compressed into summaries, embeddings, graph nodes, profiles, reflections, skills, and KV states, governance must extend from source text to all derived representations.

\paragraph{Research question.} When a memory is deleted or corrected, how can all of its derived representations---summaries, embeddings, graph nodes, materialized KV states (Section~\ref{sec:intermediate}), and learned compressed caches---be located and invalidated, and how can poisoned content be detected at write-back time at acceptable cost? Verifiable deletion and the tracking of derived representations are among the least studied stages of the memory lifecycle~\cite{lin2026memsecurity}.

\subsection{Lifecycle-Level Evaluation with Full Cost Accounting}

Current evaluation is largely based on single inputs, single answers, or fixed compression ratios, whereas memory-centric compression concerns a lifecycle in which a system must decide what to write, how to organize and retrieve it, how to compress it before injection, how to update it after inference, and what to forget. Evaluation must therefore ask not only whether a compressed system answers correctly, but whether its memory remains usable, trustworthy, and governable over time, under the stress conditions and cost reporting described in Section~\ref{sec:evaluation}.

\paragraph{Research question.} Can a single benchmark cover writing, materialization, activation, updating, and forgetting, and require methods to report the itemized costs of Section~\ref{sec:costmodel}? SCBench~\cite{li2025scbench} covers the KV lifecycle, and LongMemEval~\cite{wu2024longmemeval}, MemoryAgentBench~\cite{hu2026memoryagentbench}, and STALE~\cite{chao2026stale} cover memory recall and updating, but no benchmark evaluates both within one workload.

\subsection{Emerging Frontiers: Reasoning Traces, Multimodality, and the Parametric Boundary}

Three developments are changing the problem. In reasoning models, long chains of thought shift the main source of active memory from input to the model's own generations, and it is not yet understood which reasoning steps can be compressed without breaking correctness. Multimodal long contexts from video, document images, and embodied interaction produce far more tokens than text and have different redundancy structures, which affects both compression and cross-modal memory writing and retrieval. At the parametric boundary, test-time training, on-demand adapters, and learned compressed KV caches make the choice between holding memory in context or in weights no longer binary.

\paragraph{Research question.} Under what query frequencies, corpus sizes, and update rates should information be kept as text, materialized as KV states, or compressed into parameters? The cost model of this survey offers a starting point, but the training and update costs of parametric representations must be brought into the same account.

\section{Conclusion}
\label{sec:conclusion}

Long-context compression is evolving from a problem of length reduction into a problem of managing information for inference-time computation: the bottleneck is no longer how much input a model can accept, but which information should be retained, when it should be activated, in what form it should participate in inference, and when it should be updated or forgotten. This survey addressed this problem from a memory-centric perspective. We distinguished compute-active from compute-inactive memory, characterized intermediate states on the activation path, and described long-context compression as a lifecycle of state transitions with a budgeted cost model. Within this framework, prompt compression, latent memory, KV-cache compression, architecture-level selective computation, serving-level cache management, external long-term memory, retrieval augmentation, agent memory, and learned memory policies regulate storage cost, computational cost, and task utility at different points of one lifecycle.

Future systems need to move from static, pointwise, and modular compression toward dynamic, joint, and governable memory management: allocating memory across tiers according to task and budget, preventing errors from accumulating through write-back, supporting verifiable forgetting, and being evaluated over the full lifecycle with complete cost accounting.

\bibliographystyle{plainnat}
\bibliography{references}

\end{document}